\documentclass[10pt,twocolumn,letterpaper]{article}

\usepackage{iccv}
\usepackage{times}
\usepackage{epsfig}
\usepackage{graphicx}
\usepackage{amsmath}
\usepackage{amssymb}
\usepackage{bm}
\usepackage{bbm}
\usepackage{array}
\usepackage{multirow}
\newcommand{\vet}[1]{\mathbf{#1}}
\newcolumntype{C}[1]{>{\centering\arraybackslash\hspace{0pt}}p{#1}}
\usepackage[pagebackref=true,breaklinks=true,letterpaper=true,colorlinks,bookmarks=false]{hyperref}

 \iccvfinalcopy 

\def\iccvPaperID{36} 

\ificcvfinal\fi
\begin{document}

\title{Intra-Camera Supervised Person Re-Identification:
A New Benchmark}

\author{Xiangping Zhu$^1$, Xiatian Zhu$^2$, Minxian Li$^3$, Vittorio Murino$^{1,4}$ and Shaogang Gong$^3$\\
\\
$^1$Pattern Analysis and Computer Vision (PAVIS), Istituto Italiano di Tecnologia, $^2$Vision Semantics Ltd. \\
$^3$Queen Mary University of London, $^4$ Department of Computer Science, University of Verona\\
 {\tt\small \{xiangping.zhu2010, eddy.zhuxt\}@gmail.com, vittorio.murino@iit.it, \{m.li, s.gong\}@qmul.ac.uk}
}

\maketitle

\begin{abstract}
Existing person re-identification (re-id) methods rely mostly on a large set of 
inter-camera identity labelled training data, 
requiring a tedious data collection and annotation process therefore leading to poor scalability in practical re-id
applications.
To overcome this fundamental limitation, 
we consider person re-identification 
{\em without} inter-camera identity association
but only with
identity labels {\em independently annotated} within each individual camera-view.
%
%
This eliminates the most time-consuming and tedious inter-camera identity labelling process
in order to significantly reduce the amount of human efforts required during annotation.
It hence gives rise to a more scalable and more feasible
learning scenario, which we call {\em Intra-Camera Supervised (ICS)} person re-id.
Under this ICS setting with weaker label supervision, we formulate a Multi-Task Multi-Label (MTML) deep learning method.
Given no inter-camera association, 
MTML is specially designed for self-discovering the inter-camera identity correspondence.
This is achieved by
inter-camera multi-label learning under a joint multi-task inference framework.
In addition, MTML can also efficiently learn the discriminative re-id feature representations by fully using the available identity labels within each camera-view.
Extensive experiments demonstrate the performance superiority of our
MTML model over the state-of-the-art alternative methods on three large-scale person re-id datasets in the proposed intra-camera supervised learning setting.
\end{abstract}

\begin{figure}[t]
\centering
\includegraphics[width=.40\textwidth]{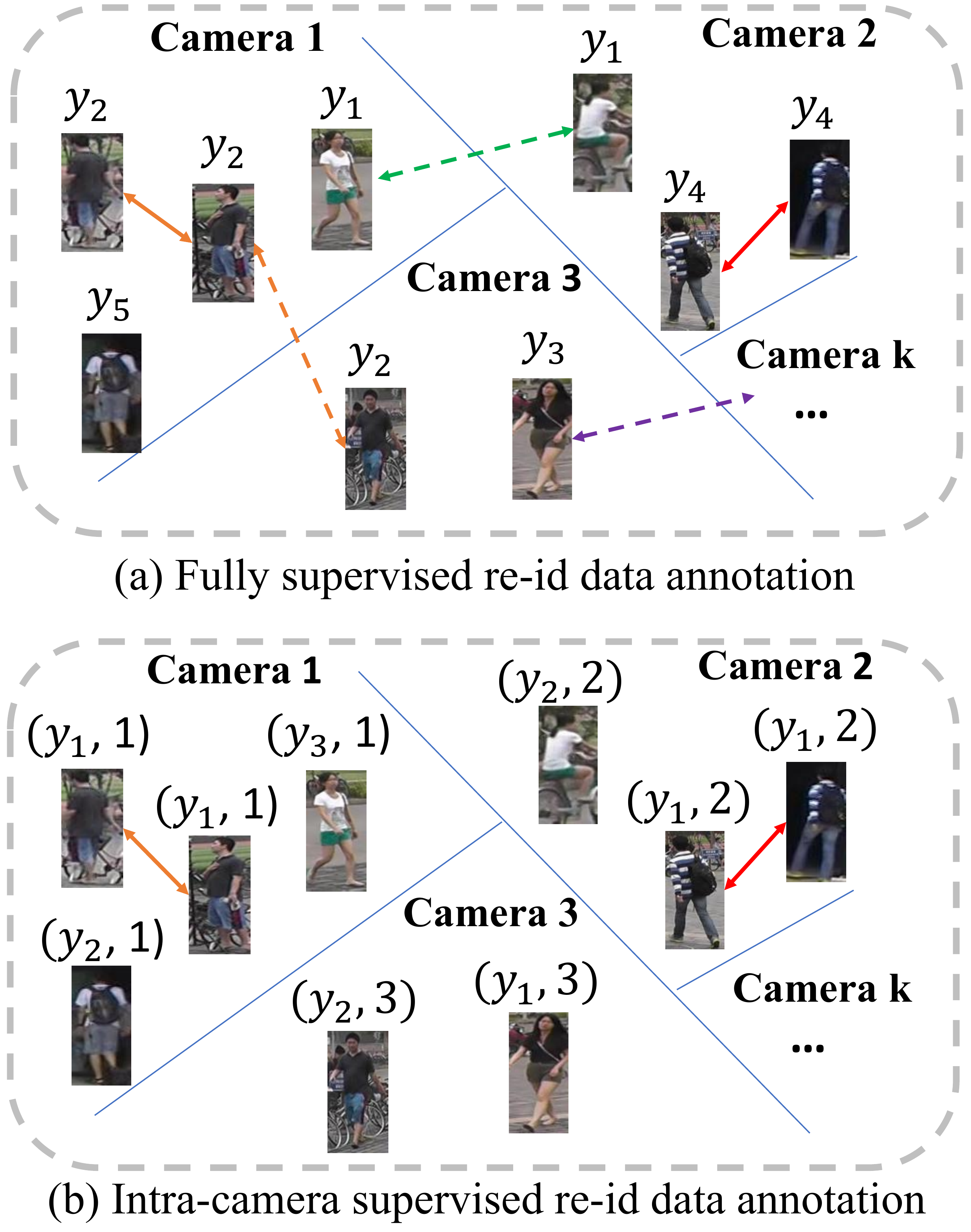}
\caption{Illustration of person re-id training data annotation:
(a) Fully supervised re-id data annotation with a single inter-camera identity label space. 
$y_i$ denotes a person ID label across cameras;
(b) The proposed intra-camera supervised re-id data annotation only with multiple camera-specific identity label spaces.
$(y_i, p)$ denotes a person ID $y_i$ with
the associated camera index $p$, i.e., per-camera independent labelling, and here, the same identity label $y_i$ in different camera-views most probably refers to different person. 
The solid and dashed arrows denote the intra-camera and inter-camera annotations, respectively.
Identity is color coded.  
For simplicity, we use one image in both (a) and (b) to denote one identity.
Compared with fully supervised re-id labeling, the inter-camera association labels are not annotated in the proposed re-id data labeling.}
\label{fig:labelling}
 \vspace{-0.5cm}
\end{figure}

\section{Introduction}
\label{sec:intro}
Person re-identification (re-id) is a task of
reasoning the subtle identity class information 
in detected person bounding box images 
captured under non-overlapping camera views
\cite{gray2008viewpoint,prosser2010person,zheng2013reidentification,market1501,XQDA,gong2014person, zhu2017exploiting}.
This is still a rather challenging task due to the nonrigid structure of the human body, the highly variable illumination conditions, and low resolution person bounding box images.
Most existing deep learning re-id methods in the literature train convolutional neural network (ConvNet) models in a supervised learning fashion \cite{cuhk03,xiao2016learning,chen2017person,zhang2018deep,li2018harmonious,chang2018multi,qian2018pose,sun2018beyond}.
One of the major limitations with supervised modelling is rooted in assuming the availability
of a large set of inter-camera labelled training 
identity classes collected through an exhaustive
and expensive annotation process.
This dramatically degrades the usability and scalability of such methods in real-world application and deployments at scales.

This problem has received a significant amount of attentions recently.
One intuitive approach is unsupervised
person re-id.
Existing methods of this kind
can be generally divided into three categories:
(1) Domain generic feature design \cite{gray2008viewpoint,farenzena2010person,market1501,XQDA,gog};
(2) Unsupervised domain adaptation
\cite{peng2016unsupervised,deng2018image,wang2018Transfer,lin2018multi,yu2018unsupervised, yu2019unsupervised, zhong2019invariance, zhu2019icip, panda2017unsupervised};
(3) Unsupervised learning
\cite{wang2016towards,chen2018deep,li2018taudl,lin2019BUC,li2019utal_pami}.
By hand-crafting universal
person appearance features, the first category aims to improve the re-id model performance generically.
However, the methods in this category often yield dramatically inferior model generalisation capability due to limited information involved in such representations.
The second category attempts to transfer the identity knowledge of a labelled source domain to
an unlabelled target domain via image or feature adaptation.
Unfortunately, such methods implicitly assume
that the source and target domains have reasonably similar camera viewing conditions for ensuring sufficient transferable knowledge.
As a more scalable approach,
the third category instead leverages only unlabelled target domain data during model training.
To benefit from existing supervised learning algorithms, previous unsupervised re-id methods 
often turn to the idea of self-discovering the underlying identity association information.
Compared with conventional manual labelling,
this automated annotating remains 
less accurate and less complete, 
leading to inferior model optimisation and generalisation capability.

In this work, we instead investigate the person re-id scalability from the data annotation perspective. 
We consider a more scalable re-id problem with cheaper
training data labelling where person identity (ID) labels are 
annotated in each camera-view {\em without} inter-camera association.
This is based on our observation that inter-camera search in the manual annotation is the most time-consuming and expensive sub-process (Fig \ref{fig:labelling}{\color{red} a}).
It is because, the generic (unframed) people usually takes a-prior unknown routines in open public space with complex space-time topology.
On the other hand, labelling person identity classes
in each camera view {\em independently} is much simpler and faster,
possibly further benefiting from
off-the-shelf tracking algorithms in a single camera view
(Fig \ref{fig:labelling}{\color{red} b}).
We name this new setting as
{\em Intra-Camera Supervised} (ICS) person re-id.
Compared with conventional strong re-id supervision
with labelled inter-camera identity association,
this re-id problem focuses a learning algorithm
on self-discovering the correspondence relationships between
camera-specific identity spaces. It presents a new modelling challenge.

To address the ICS re-id problem, we proposed 
a Multi-Task Multi-Label (MTML) deep learning model in this work. 
Since there is no inter-camera association in the proposed re-id data labeling, MTML is specially designed for self-discovering the inter-camera identity correspondence by the 
inter-camera multi-label learning component under a joint multi-task inference framework.
Some previous inter-camera identity association re-id methods \cite{chen2018deep,li2018taudl,li2019utal_pami}
learn discriminative representations by associating similar samples in the feature space.
In contrast, we introduce an idea of multiple labels for each person
identity in inter-camera association across different label spaces
for better exploiting the per-camera identity labelling
information.
In addition, MTML can also efficiently learn the discriminative re-id features using the provided per-camera identity labels based on multi-camera multi-task learning.

The {\bf contributions} of this work are:
{\bf (1)} We reformulate supervised learning person re-id by removing explicitly the assumption for exhaustive
inter-camera pairwise labelling in model training.
This eliminates the most time-consuming
and tedious inter-camera pairwise ID labelling task required in the
conventional re-id model training. 
In return, a more challenging re-id model learning problem is presented with only per-camera independently annotated ID labels in the training dataset.
Compared to completely unsupervised re-id, our introduced re-id problem enables the re-id model to benefit from per-camera view labelled ID information which can be easily annotated or generated using tracking algorithms.
{\bf (2)} We formulate a Multi-Task Multi-Label (MTML) learning method for ICS person re-id.
MTML takes a multi-task learning framework to jointly account for independent camera-specific 
identity discriminative labelling information and self-discovering inter-camera identity association in a multi-labelling fashion. 
Three large-scale re-id datasets, i.e., Market-1501 \cite{market1501}, 
DukeMTMC-reID \cite{DukeMTMC-reID, ristani2016MTMC}, and MSMT17 \cite{wei2018person}, have been used in experiments with the proposed ICS setting. The results demonstrate the superiority of the proposed MTML method compared with the state-of-the-art person re-id models.

\section{Related Work}
\label{sec:related_works}
As we are concerned with the re-id scalability issue
from the person re-id dataset annotation perspective,
this section will discuss and review supervised and unsupervised
person re-id works in the literature.

Supervised learning based person re-id methods dominate
the literature \cite{xiao2016learning,wei2018person,chen2017person,wang2018reg,zhao2017deeply,zhu2017fast,chen2017tpami,li2017person,chang2018multi,zhang2018deep,Shen_2018_ECCV,li2018harmonious,qian2018pose,sun2018beyond}.
This type of models are trained in a {\em strongly} supervised manner
by inter-camera pairwise ID labelled training images. 
They suffer from significant model performance degradation when the test domain is dissimilar
to the training domain.
Moreover, supervised learning based re-id models are effective only when strongly labelled
training data are available at large scale for every target domain. This
limits their usefulness. Semi-supervised learning methods
\cite{liu2014semi,wang2016towards}
decrease the amount of labelled training data but still require some
cross-view pairwise labelling. Removing the expensive inter-camera
pairwise labelling requirement for re-id model learning is desirable in practice.

Unsupervised learning based person re-id models have received
increasing attention with three flavours, i.e.,
domain generic feature design \cite{gray2008viewpoint,farenzena2010person,market1501,XQDA,gog}, unsupervised domain adaptation
\cite{wang2018Transfer,lin2018multi,zhong2018generalizing,yu2018unsupervised, zhong2019invariance}
 and unsupervised model learning
\cite{lisanti2015person,wang2016towards,kodirov2016person,ma2017person,chen2018deep,li2018taudl,lin2019BUC,li2019utal_pami}.
All these methods do not need labelled training data from the target domain
therefore more deployable.
However, their re-id performances are much weaker than those of
supervised learning based models (when training and test domains are
similar).

Intra-camera supervised learning based person re-id is
considered in this work, where the strong inter-camera identity association labels are removed
from the training data. 
Without the need for manually annotating identity 
correspondences between every pair of camera views,
we minimise the amount of labours required
for person identity class annotation and enable a re-id model to be more
feasible in deployment at scale.
To solve this re-id problem, we develop
a re-id learning algorithm for 
making full use of per-camera independently annotated ID labels
and self-discovering most
likely person correspondences between different camera views,
yielding stronger re-id models than the unsupervised
learning counterpart. 

\section{Methodology}
\label{sec:reid_model}

In this section we formulate a person re-id learning method without
inter-camera identity association in the training data.
Suppose there are $M$ camera views in a camera network. 
For the $p$-th camera view, we {\em independently} annotate a set of samples
$\mathcal{D}_p = \{(\vet{x}_i, y_k, p)\}$
where $\vet{x}_i$ is the $i$-th person image in $\mathcal{D}_p$. Each person image $\vet{x}_i$ is associated with an identity label $y_k\in \{y_1,y_2,\cdots,y_{N_p}\}$ and the corresponding camera identity $p \in \{ 1, 2, \cdots, M\}$. $N_p$ is the total number of unique person identities in $\mathcal{D}_p$.
Due to per-camera independent labelling nature,
the same identity labels (e.g. identity $y_1$) of any two camera views are very likely referring to two different persons.
By combining camera-specific labelled data,
we obtain the entire training set as
$\mathcal{D} = \{\mathcal{D}_{1}, \mathcal{D}_{2}, \ldots, \mathcal{D}_{M}\}$.
%
The presence of such multiple identity class label spaces
prevents the training of a conventional supervised re-id model and thus a new effective re-id method is needed.

\begin{figure*}[t]
\centering
\includegraphics[width=0.99 \textwidth]{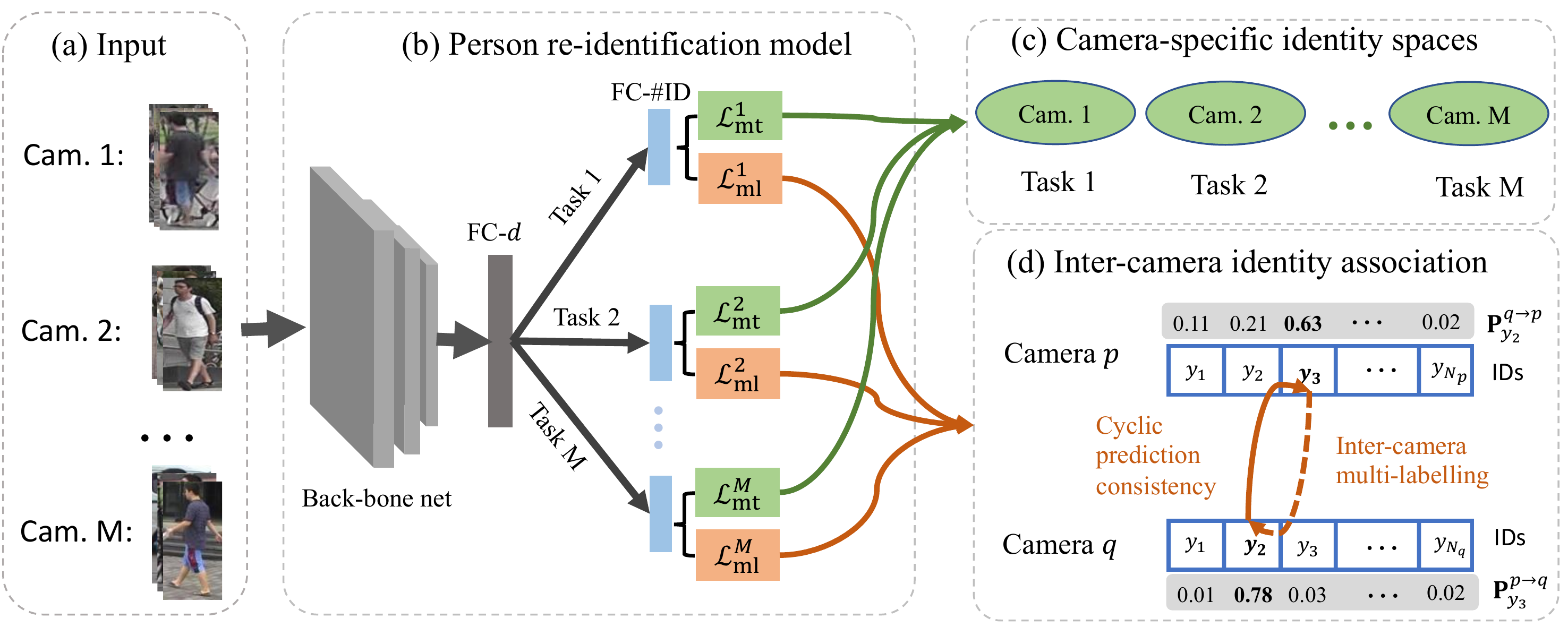}
\caption{Overview of the proposed Multi-Task Multi-Label (MTML) deep learning method for intra-camera supervised person re-id.
{\bf (a)} Given input person images which are labeled independently in each camera view,
MTML aims to derive {\bf (b)} an identity discriminative feature representation.
This is achieved by 
designing two components:
(1) {\em multi-camera multi-task learning}
where each individual camera view is assigned
with {\bf (c)} 
a separate supervised learning task for modelling the corresponding identity space,
and (2) {\em inter-camera multi-label learning}
where {\bf (d)} inter-camera identity association labels
are self-discovered using the cyclic prediction consistency strategy. These association labels are further included into the training dataset to impose the re-id model to modeling the the inter-camera person appearance variation under the multi-task inference framework. 
The re-id model is composed with the backbone network and fully connected layers, i.e., FC-$d$ and FC-\#ID, in which $d$ denotes the dimension of the feature representation and \#ID denotes the number of identities in corresponding camera-view.
}
\label{fig:pipeline}
\end{figure*}

We formulate a Multi-Task Multi-Label (MTML) deep learning method
for addressing this more challenging and more scalable ICS re-id problem.
Given the per-camera identity independent labelling nature, the key of model learning lies in two aspects: 
(1) How to effectively exploit the per-camera identity labels,
and (2) How to associate the identity label classes across camera views (or label spaces).
MTML achieves these two aspects by integrating two corresponding components:
(i) Multi-camera multi-task learning
that assigns a separate learning task to each individual
camera view for modelling the respective identity space, 
(ii) inter-camera Multi-label learning
that automatically self-discovers the identity associations
across camera views and assigns multiple labels to these associated identities for modeling the inter-camera person appearance variation.
More details about these two components are presented in the following parts and Fig. \ref{fig:pipeline} gives an overview of the proposed MTML model.

\subsection{Multi-Camera Multi-Task Learning}
\label{subsec:mcmt}
As shown in Fig. \ref{fig:pipeline}{\color{red} b} and \ref{fig:pipeline}{\color{red} c}, we consider the multi-task learning strategy 
\cite{argyriou2007multi} given camera independently labelled
person identity information.
This aims at better mining the common knowledge 
shared across camera views whilst enhancing model learning by augmented training data for each camera view.
Each camera view is treated separately due to their
independent labelling property.
Importantly, this also allows to derive a person re-id representation with implicit inter-camera identity discriminative capability for facilitating inter-camera identity association
\cite{li2019utal_pami}.

Formally, we create a camera-shared feature representation
upon which multi-task branches
are rooted.
Each branch is responsible for the classification task in a specific camera view.
During model training, each task branch can be used to propagate the 
respective per-camera identity label information
via the softmax cross-entropy loss function. For one sample $(\vet{x}_i, y_k, p)\in \mathcal{D}_p$, its corresponding softmax cross-entropy loss function can be formulated as:
\begin{equation}
\mathcal{L}_{\text{mt}, i}^p =
{-} 
\mathbbm{1} (y_k)
{\log} f_p(\vet{v}_i)
\label{eq:mt_loss_one_sample}
\end{equation}
where $\vet{v}_i\in \mathbbm{R}^{d\times 1}$ specifies the {\em camera-shared} feature vector of the corresponding image $\vet{x}_i$ from the $p$-th camera and it is extracted after the fully connected layer FC-$d$ as shown in Fig. \ref{fig:pipeline}, in which $d$ is the dimension of a feature vector. $f_p(\cdot): \mathbbm{R}^{d\times 1} \to \mathbbm{R}^{N_p\times 1}$ denotes the classifier function of camera $p$. 
The one-hot encoding function $\mathbbm{1} (\cdot): \mathbbm{R} \to \mathbbm{R}^{1\times N_p}$ returns an one-hot vector with value 1 for the element at the given index.
For stochastic mini-batch deep learning,
the {\em  multi-camera multi-task learning} (MT) objective 
is designed as:
\begin{equation}
\mathcal{L}_\text{mt}= \frac{1}{B} \sum_{p=1}^{M}
\mathcal{L}_\text{mt}^p
\label{eq:mt_loss_one_batch}
\end{equation}
where $\mathcal{L}_\text{mt}^p$ denotes the accumulated cross-entropy loss
(Eq. \eqref{eq:mt_loss_one_sample}) of all in-batch images from the $p$-th camera, and $B$ is the mini-batch size. With this multi-camera multi-task learning, the discriminative re-id features can be efficiently learned using the existing identity labels within each camera-view.

\subsection{Inter-Camera Multi-Label Learning}
\label{subsec:cross_cam_learning}

In person re-id, inter-camera person appearance variation
is one of the most significant elements
during model training.
Whilst this is implicitly learned in the multi-camera 
multi-task learning component as discussed above,
it is insufficient to fully capture the underlying
inter-camera identity correspondence relationships.
To address this problem, an inter-camera multi-label
learning component is designed that aims to self-discover the identity correspondence between camera-specific identity label spaces and imposes the re-id model to effectively model the inter-camera person appearance variation.

Specifically, given an identity class $y_k\in \{y_1, y_2, \cdots,$ $y_{N_p}\}$ from camera
$p \in \{1, 2, \ldots, M\}$,
we want to find if a true match exists in camera $q$.
To this end, all the person images
of $y_i$ are mapped into the branch of camera $q$ and an average
prediction of $y_i$ in camera $q$ is obtained as:
\begin{align}
  \vet{P}^{p \to q}_{y_k} = \text{avg} ({f}_q(\vet{v}_i)),
  \label{eq:cross_cam_mapping}
\end{align}
in which $\vet{P}^{p \to q}_{y_k}\in \mathbbm{R}^{N_q\times 1}$ is the averaged prediction of $y_k$ in camera $q$. As in Eq. \eqref{eq:mt_loss_one_sample}, $f_q(\cdot)$ denotes the classifier function of camera $q$
and $\text{avg}(\cdot)$ is the averaging function.
We then nominate the identity class in camera $q$
with the maximum likelihood probability as the candidate matching identity:
\begin{align}
    l^{*} = \arg \max_{l \in \{ 1, 2, \ldots, N_q\}} 
    \vet{P}^{p\to q}_{y_k} (l),
    \label{eq:cross_cam_match}
\end{align}
where $l^*$ is the index of identity $y_{l^*}\in \{y_1, y_2, \cdots, y_{N_q} \}$ in $q$-th camera.
To boost the accuracy and robustness of matching pairs, the identity $y_{l^*}$ is further mapped back to camera $p$
in a similar way as Eq. \eqref{eq:cross_cam_mapping}
and the corresponding 
candidate matching identity $y_{t^*}$ as in Eq. \eqref{eq:cross_cam_match} is retrieved.
This cyclic mapping and matching operation
between every two camera views determines the 
inter-camera identity association result as:
\begin{equation}
\left\{ 
    \begin{array}{ll}
      (y_k, y_{l^*}) \;\; \text{is a matching pair}, & \text{if} \;\; y_{t^*} = y_k, \\
      (y_k, y_{l^*}) \;\; \text{is a unmatching pair}, & \text{otherwise}.
    \end{array}
  \right.   
\label{eq:match_pair}
\end{equation}

To benefit model training from self-discovered identity matching pairs, a proper supervision function is designed.
Considering the idea of inter-camera prediction based identity association,
the inter-camera learning is performed by multi-labeling the associated person identities.

In particular, given an identity matching pair ($y_k,y_{l^*}$),
with $y_k$ from camera $p$ and $y_{l^*}$ from camera $q$ as defined
in Eq. \eqref{eq:match_pair},
the person images corresponding to the associated matching identities $y_k$ and $y_{l^*}$ are assigned with multiple labels. 
For the person images of $y_k$, they are assigned with the new label $(y_{l^*}, q)$ and similarly, 
the person images of $y_{l^*}$ are assigned with the new label $(y_k, p)$. 
After this inter-camera multi-labeling, the images of $y_k$ and $y_{l^*}$ are attached with the identical multiple identity labels, 
i.e., $(y_k, p)$ and $(y_{l^*}, q)$, and thus $y_k$ and $y_{l^*}$ are inter-camera associated.

With these newly assigned labels, a simple but efficient loss function is designed based on the softmax cross entropy loss as the MT loss in Eqs. \eqref{eq:mt_loss_one_sample} and \eqref{eq:mt_loss_one_batch}. For one person image $\vet{x}_i$ with new label $(y_{l^*}, q)$ as in Eq. \eqref{eq:cross_cam_match}, its ML loss can be formulated as:
\begin{equation}
\mathcal{L}_{\text{ml},i}^q =
{-} \mathbbm{1} (y_{l^*})
{\log} f_q(\vet{v}_i)
\label{eq:one_sample_ml_loss}
\end{equation}
in which $\vet{v}_i$ is the feature vector corresponding to the person image $\vet{x}_i$. The definitions of $\mathbbm{1} (\cdot)$ and $f_q(\cdot)$ are the same as in Eq. \eqref{eq:mt_loss_one_sample}. As the MT loss in Eq. \eqref{eq:mt_loss_one_batch}, the final ML loss for one mini-batch is defined as:
\begin{equation}
\mathcal{L}_\text{ml}= \frac{1}{M} \sum_{q=1}^{M}
\mathcal{L}_\text{ml}^q
\label{eq:ml_loss}
\end{equation}
in which $\mathcal{L}_\text{ml}^q$ is the ML loss in camera $q$, i.e., $\mathcal{L}_\text{ml}^q=\frac{1}{b_q}\sum_{i=1}^{b_q} \mathcal{L}_{\text{ml},i}^q$. $b_q$ is the number of images with newly assigned labels in camera $q$.


\subsection{Model Objective Loss Function}
By combining the multi-camera and multi-label learning
function in a multi-task manner, we 
obtain the final MTML model objective loss function as:
\begin{align}
\mathcal{L} = \mathcal{L}_\text{mt} + \lambda \mathcal{L}_\text{ml},
\label{eq:final_objective_loss}
\end{align}
where $\lambda$ controls the relative weight of the two terms. In our experiments, we set $\lambda=0.5$ considering that
inter-camera identity association is necessarily noisy therefore
adversely affects the quality of $\mathcal{L}_\text{ml}$.

\subsection{Model Training and Inference}
The stochastic gradient descent algorithm
can be applied for optimizing the proposed deep re-id model.
In the considered re-id dataset annotation as shown in Fig. \ref{fig:labelling}(b), per-camera identity labels are accurately annotated
whilst the identity matching between camera views is likely inaccurate.
Based on this observation, the proposed re-id model is first pre-trained using 
only the multi-camera multi-task learning loss $\mathcal{L}_{\text{mt}}$ (Eq \eqref{eq:mt_loss_one_batch}). 
Then based on this pre-trained re-id model, the inter-camera multi-label learning is iteratively performed using the model objective loss function $\mathcal{L}$ as in Eq. \eqref{eq:final_objective_loss}.
In every iteration, the re-id model will be first trained for a number of epochs, and then the cyclic prediction consistency and inter-camera multi-labeling will be applied for associating inter-camera identities. 
The newly assigned multi-labels of associated identities will be further included into the training dataset for model learning in the following iteration.

In model inference, the trained MTML model is deployed
to extract the camera-shared features of test person images
as their re-id representations.
For efficient re-id matching and ranking, the Euclidean distance metric is 
utilised to compute the probe-gallery pairwise similarity.

\begin{table}[t]
	\center
	\caption{Comparisons between the proposed MTML method and existing re-id methods on Market1501 dataset.}
	\vspace{0.2cm}
	\label{tb:market_sota}
	\begin{tabular}{c||ccc|c}
		\hline
		\hline
		Metric 
		& R1 & R10 & R20 & mAP\\
		\hline \hline
		PUL \cite{fan18unsupervisedreid}
		& {41.9} & {64.3} & {70.5} & {18.0}\\
		CAMEL \cite{yu2017cross} 
		&  {54.5} & - & - & {26.3} \\
		TJ-AIDL \cite{wang2018Transfer} 
		& {58.2} & {-} & {-} & {26.5}\\
		CycleGAN \cite{zhu2017unpaired} 
		& {48.1} & {-} & {-} & {20.7}\\
		SPGAN \cite{deng2018image} 
		& {51.5} & {76.8} & {82.4} & {22.8}\\
		SPGAN+LMP \cite{deng2018image} 
		& {58.1} & {82.7} & {87.9} & {26.9}\\
		HHL \cite{zhong2018generalizing} 
		& {62.2} & {84.0} & {88.3} & {31.4}\\
		MAR \cite{yu2019unsupervised} 
		& {67.7} & {-} & {-} & {40.0}\\
		ECN \cite{zhong2019invariance} 
		& {75.1} & {91.6} & {-} & {43.0}\\
		\hline
		E-PCSL
		& {42.6} & 64.6 & 69.9 & {17.6}  \\
		UTAL \cite{li2019utal_pami}                
		& {69.2} & 85.5 & 89.7 & {46.2} \\
		\hline
		\bf MTML 
		& \bf{85.3} & \bf{96.2} & \bf{97.6} &\bf {65.2} \\
		\hline
		\hline
	\end{tabular}
\end{table}

\section{Experiment}
\label{sec:exp}

\subsection{Experimental Setup}
\noindent {\bf Datasets.}
Three large-scale re-id datasets, i.e.,
Market-1501 \cite{market1501}, 
DukeMTMC-reID \cite{DukeMTMC-reID,ristani2016MTMC}, and MSMT17 \cite{wei2018person}, are selected for evaluating our proposed ICS problem and our MTML method.
As no existing re-id datasets annotated in the ICS fashion, we adopted these three fully labelled
re-id datasets by independently annotating their identity labels in each camera-view as shown in Fig. \ref{fig:labelling}.
%
%
We still utilise the identical test data
of each dataset for model performance evaluation.
We will publicly release these ICS person re-id benchmarks.

\vspace{0.1cm}
\noindent {\bf Performance metrics.}
We used the common Cumulative
Matching Characteristic (CMC) and mean Average Precision (mAP) metrics for model performance measurement.

\vspace{0.1cm}
\noindent{\bf Implementation details.}
In practice, the ImageNet pre-trained
ResNet-50 \cite{resnet} is selected as the backbone CNN \footnote{Layers after avg-pooling are removed.} of our MTML model. For multi-task learning, each 
branch is formed by a FC classification layer. 
Person bounding box images are resized to 256$\times$128
in pixel before feeding into the network.
The standard stochastic gradient descent (SGD) optimizer is adopted for training the MTMC model with the initial learning rate of 0.05. 
In pre-training the model using only the MT loss, the learning rate is decayed 10 times every 40 epochs and the epoch number is set to 100.
In inter-camera multi-label learning, the learning rate is decayed 10 times after 8 epochs. The epoch number is set to 15 in each iteration and the number of iteration is set to 8.
In order to balance
the model training speed across camera views, we randomly
selected from each camera view the same number of images (i.e., 4 images) from one person identity and the same number of persons (i.e., 2 persons).
By default, we set $\lambda=0.5$ in Eq \eqref{eq:final_objective_loss} for balancing the the losses of $\mathcal{L}_\text{mt}$ and $\mathcal{L}_\text{ml}$.

\begin{table}[t]
	\center
	\caption{Comparisons between the proposed MTML method and existing re-id methods on the DukeMTMC-reID dataset.}
	\vspace{0.2cm}
	\label{tb:duke_sota}
    \begin{tabular}{c||ccc|c}
		\hline
		\hline
		Metric 
		& R1 & R10 & R20 & mAP\\
		\hline \hline
		PUL \cite{fan18unsupervisedreid}
		& {23.0} & {39.5} & {44.2} & {12.0}\\
		TJ-AIDL \cite{wang2018Transfer} 
		& {44.3} & {-} & {-} & {23.0}\\
		CycleGAN \cite{zhu2017unpaired} 
		& {38.5} & {-} & {-} & {19.9}\\
		SPGAN \cite{deng2018image} 
		& {41.1} & {63.0} & {69.6} & {22.3}\\
		SPGAN+LMP \cite{deng2018image} 
		& {46.9} & {68.5} & {74.0} & {26.4}\\
		HHL \cite{zhong2018generalizing} 
		& {46.9} & {66.7} & {71.9} & {27.2}\\
		MAR \cite{yu2019unsupervised} 
		& {67.1} & {-} & {-} & {48.0}\\
		ECN \cite{zhong2019invariance} 
		& {63.3} & {80.4} & {-} & {40.4}\\
		\hline
		E-PCSL
		& {38.8} & 58.9 & 64.6  & {22.1}  \\
		UTAL \cite{li2019utal_pami}                
		& {62.3} & 80.7 & 84.4 & {44.6} \\
		\hline
		\bf MTML              
		& \bf {71.7} & \bf{86.9} & \bf{89.6} &\bf {50.7} \\
		\hline
		\hline
	\end{tabular}
\end{table}

\begin{table}[t]
	\center
	\caption{Comparisons between the proposed MTML method and existing re-id methods on the MSMT17 dataset.}
	\vspace{0.2cm}
	\label{tb:msmt_sota}
    \begin{tabular}{C{2.0cm}||C{0.9cm}C{0.9cm}C{0.9cm}|C{0.9cm}}
		\hline
		\hline
		Metric 
		& R1 & R10 & R20 & mAP\\
		\hline
		\hline
		PTGAN \cite{wei2018person}
		& {11.8} & {27.4} & {-} & {3.3}\\
		ECN \cite{zhong2019invariance}
		& {30.2} & {46.8} & {-} & {10.2}\\
		\hline
		E-PCSL
		& {16.8} & 31.5 & 37.4 & {5.4}  \\
		UTAL \cite{li2019utal_pami}                
		& {31.4} & 51.0 & 58.1  & {13.1} \\
		\hline
		\bf MTML              
		&\bf {44.1} & \bf{63.9} & \bf{70.0} &\bf {18.6} \\
		\hline
		\hline
	\end{tabular}
\end{table}

\subsection{Evaluation on Person Re-Identification}
\label{subsec:baselines}

\noindent {\bf Evaluated methods.} Apart from the proposed MTML model, we further evaluated two methods particularly adapted to the newly introduced ICS person re-id setting:
{\bf (1)} {\em Ensemble of Per-Camera Supervised Learning} (E-PCSL):
Without inter-camera ID labels, we trained a separate re-id model
for each camera on the corresponding labelled training data.
We used the ResNet-50 as the backbone CNN,
and the softmax cross-entropy loss function as 
the supervised objective.
During deployment, given a test image we extracted
the feature vectors of all the per-camera models,
concatenated them into a single representation vector,
and utilised the Euclidean distance for re-id matching.
{\bf (2)} {\em Unsupervised Tracklet Association Learning} (UTAL) \cite{li2019utal_pami}:
This method is designed for associating person tracklets in an unsupervised manner 
for video based re-id, taking the auto-detected tracklets as imagery data form in particular. 
For enabling multi-shot image based re-id as considered here, 
following UTAL we stacked the images with the same ID from the same camera into a single tracklet, forming the intra-camera supervision specifically for this model. 
In terms of experiment setting, UTAL assumes the same training data annotation as in our work. 
However, it is noteworthy to mention that this is due to lacking spatial-temporal information 
in the existing image based person re-id benchmarks; Conceptually, the two works
investigate rather different person re-id scenarios, starting from 
distinctive motivations and annotation assumptions.
%

In addition, the proposed MTML is also compared with the state-of-the-art unsupervised domain adaptive re-id methods which consider the re-id problem with fully labelled data in the source domain but no labels in the target domain. These methods include CAMEL \cite{yu2017cross},
PUL \cite{fan18unsupervisedreid},
TJ-AIDL \cite{wang2018Transfer},
CycleGAN \cite{zhu2017unpaired},
SPGAN \cite{deng2018image},
PTGAN \cite{wei2018person},
HHL \cite{zhong2018generalizing},
MAR \cite{yu2019unsupervised}, ECN \cite{zhong2019invariance}.
This provides a overall quantitative evaluation
and comparision between different person re-id settings,
but no apple-to-apple comparison due to different types of supervision involved.

\vspace{0.2cm}
\noindent {\bf Results.}
Tables \ref{tb:market_sota}-\ref{tb:msmt_sota} give the re-id performance comparison results between our MTML model and other considered methods. 
Several observations can be derived that:
{\bf (1)} By independently exploiting camera-specific identity class annotation,
the baseline E-PCSL yields the weakest re-id model generalisation. 
This is due to the incapability of 
leveraging the shared knowledge 
between camera views and mining the inter-camera identity matching information. 
{\bf (2)} 
The model performance is continuously increased by
more recent unsupervised re-id models.
In comparison, the proposed MTML model improves the performance observably.
One reason is that our model benefits from more scalable per-camera ID labelling, in addition to the superior formulation of our model.
{\bf (3)}
The proposed MTML model significantly outperforms both E-PCSL and UTAL, suggesting the performance superiority 
of our method in tackling the person re-id problem under the proposed cheaper annotation case.
Whilst MTML shares partly the model structure with UTAL
in terms of multi-task learning design, we observed
a large performance difference between them.
The plausible reason may be due to 
the unique advantage of exploiting the cyclic prediction consistency based 
inter-camera identity association.

\begin{table} [h]
	\center
\caption{
Effectiveness analysis of two components in MTML: multi-camera multi-task (MT) and inter-camera multi-label (ML).} 
\vspace{0.2cm}
	\label{tb:abalation_study}
	\begin{tabular}{C{1.5cm}||C{1cm}C{1cm}C{1cm}|C{1cm}}
		\hline
		\hline
		dataset &\multicolumn{4}{c}{Market-1501} \\
		\hline
		metric  & R1 & R10 & R20 & mAP \\
		\hline
		MT
		& {78.4} & 93.1 & 95.7  & {52.1}\\
		\hline
		\bf MTML        
		& \bf{85.3} & \bf{96.2} & \bf{97.6} &\bf {65.2}\\
		\hline
		\hline
				dataset &\multicolumn{4}{c}{DukeMTMC-reID} \\
		\hline
		metric  & R1 & R10 & R20 & mAP \\
		\hline
		MT
		& 65.2 & 81.1 & 85.6 & 44.7 \\
		\hline
		\bf MTML        
		& \bf {71.7} & \bf{86.9} & \bf{89.6} &\bf {50.7} \\
		\hline
		\hline
				dataset &\multicolumn{4}{c}{MSMT17} \\
		\hline
		metric  & R1 & R10 & R20 & mAP \\
		\hline
		MT
		& 39.6 & 59.6 & 65.7 & 15.9\\
		\hline
		\bf MTML        
		& \bf {44.1} & \bf{63.9} & \bf{70.0} &\bf {18.6}\\
		\hline
		\hline
	\end{tabular}
\end{table}

\subsection{Further Analysis and Discussions}

\noindent {\bf Model component analysis.}
We examined the effectiveness of the two model components in MTML, i.e.,
multi-camera multi-task (MT) and inter-camera multi-label (ML) learning.
Table \ref{tb:abalation_study} shows that:
(1) With the MT component alone, 
the model can already achieve very competitive
performance, suggesting the significance of sharing labelling
knowledge across all the camera views 
via joint multi-task inference.
(2) After adding the ML component,
the model generalisation capability can be further boosted.
This indicates the positive influences of 
leveraging the inter-camera identity association information
through self-supervision despite at the risk of
deriving false inter-camera identity associations and propagating their 
error information into re-id model during training.

\begin{figure}[h]
\centering
\includegraphics[width=0.46\textwidth]{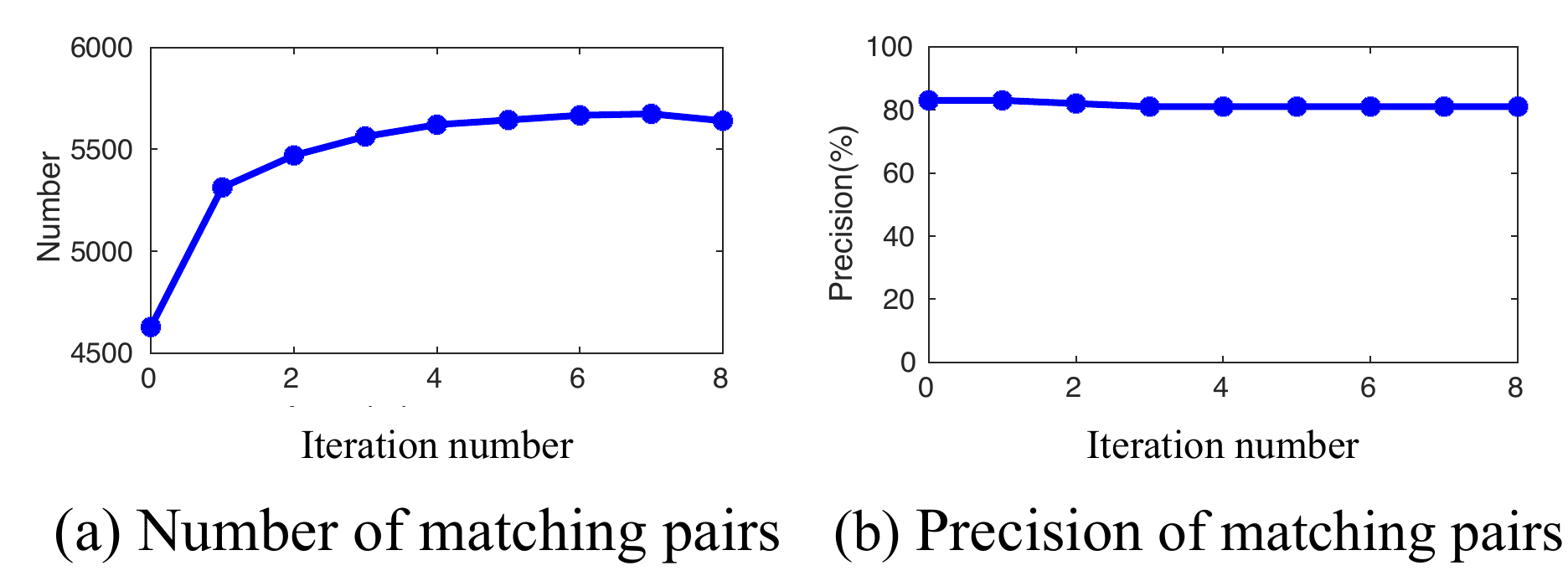}
\caption{Dynamics of self-discovered inter-camera identity
matching pairs during model training on the Market-1501 dataset.}
\label{fig:match_pair_dynamic}
\end{figure}

\vspace{0.1cm}
\noindent {\bf Inter-camera identity association dynamics.}
To further examine the benefits of inter-camera identity association,
we tracked the evolving dynamics of self-discovered matching pairs during the model training process.
Figure \ref{fig:match_pair_dynamic} shows that 
our model is able to
reveal an increasingly number of inter-camera identity matching pairs 
whilst maintaining high association accuracy. This
explicitly explains the model performance advantage of 
the proposed inter-camera multi-label learning idea.

\section{Conclusion}
\label{sec:conclusions}
In this work, we presented 
a more scalable intra-camera supervised (ICS) person re-identification problem, characterised by 
re-id model learning
without cross-view pairwise labelling 
but with only per-camera independent person
identity labels. 
The key idea is to eliminate the tedious process of
manually annotating exhaustively identity classes across every pair of
camera views in a surveillance network, both costly and sparsely available. 
This reformulates the conventional supervised re-id model learning into a weakly supervised
learning problem with
multiple independent ID label spaces across camera views.
Consequently, it focuses the
learning task on self-discovering inter-camera identity
label associations. 
To that end, we introduced a Multi-Task Multi-Label
(MTML) learning algorithm capable of fully exploiting the available
weak re-id supervision constraint whilst simultaneously self-mining inter-camera
identity association by a cyclic classification consistency idea. 
Extensive evaluations were conducted on three re-id benchmarks to validate the advantages
of the proposed MTML model over the state-of-the-art alternative methods in the proposed ICS learning setting.
The detailed component analysis is also provided for giving insights on our model design.

\section*{Acknowledgement}
{\noindent This work was partially supported by Vision Semantics Limited, the Alan 
Turing Institute Fellowship Project on Deep Learning for Large-Scale 
Video Semantic Search, and the Innovate UK Industrial Challenge Project 
on Developing and Commercialising Intelligent Video Analytics Solutions 
for Public Safety (98111-571149).}

{\small
\bibliographystyle{ieee}
\bibliography{main.bib}

\end{document}